\documentclass[journal]{IEEEtran}
\usepackage{amsmath,amsfonts}
\usepackage{algorithmic}
\usepackage{algorithm}
\usepackage{array}
\usepackage[caption=false,font=normalsize,labelfont=sf,textfont=sf]{subfig}
\usepackage{textcomp}
\usepackage{stfloats}
\usepackage{url}
\usepackage{verbatim}
\usepackage{graphicx}
\usepackage{cite}
\usepackage{booktabs} 
\usepackage{multirow} 
\usepackage{float} % 确保在导言区引入 float 宏包
\usepackage{placeins} % 防止表格漂移
\usepackage{hyperref}
\usepackage{xcolor}

\begin{document}

\title{TGT: A Temporal Gating Transformer for Smartphone App Usage Prediction}

\author{
    Longlong Li,
    Cunquan Qu,\thanks{*Corresponding author.}
    and Guanghui Wang
    
    \thanks{Longlong Li is with the School of Mathematics and the Data Science Institute, Shandong University, Jinan 250100, China (e-mail: longlee@mail.sdu.edu.cn).}
    
    \thanks{*Cunquan Qu is the corresponding author and is with the Data Science Institute, Shandong University, Jinan 250100, China (e-mail: cqqu@sdu.edu.cn).}

    \thanks{Guanghui Wang is with the School of Mathematics, Shandong University, Jinan 250100, China (e-mail: ghwang@sdu.edu.cn).}
}

\maketitle
\begin{abstract}
Accurately predicting smartphone app usage is challenging due to the sparsity and irregularity of user behavior, especially under cold-start and low-activity conditions. Existing approaches mostly rely on static or attention-only architectures, which struggle to model fine-grained temporal dynamics. We propose TGT, a Transformer framework equipped with a temporal gating module that conditions hidden representations on the hour-of-day. Unlike conventional time embeddings, temporal gating adaptively rescales feature dimensions in a time-aware manner, working orthogonally to self-attention and strengthening temporal sensitivity. TGT further incorporates a context-aware encoder that integrates session sequences and user profiles into a unified representation. Experiments on two real-world datasets, Tsinghua App Usage and LSApp, demonstrate that TGT significantly outperforms 15 competitive baselines, achieving notable gains in HR@1 and maintaining robustness under cold-start scenarios. Beyond accuracy, analysis of gating vectors uncovers interpretable daily usage rhythms, showing that TGT learns human-consistent patterns of app behavior. These results establish TGT as both a powerful and interpretable framework for time-aware app usage prediction.
\end{abstract}


\section{Introduction}

With the ubiquity of smartphones, mobile apps have become essential to daily life—shaping how people communicate, work, and access services~\cite{zhu2013mobile, mihailidis2014tethered, chen2019cap, tu2019fingerprint, zhang2020app,AKDIM2022102888}. Accurately predicting user app usage is vital for enabling personalized recommendations~\cite{liu2017cm, zhao2019user}, optimizing device performance~\cite{chen2017powerful, oliner2013carat}, and assisting network-level resource scheduling~\cite{xu2016understanding, zeng2018temporal}. However, app usage prediction remains challenging due to inherent characteristics of user behavior: high dimensionality, temporal sparsity, and complex, time-dependent patterns~\cite{pejovic2015anticipatory, han2016mobile}. These challenges are particularly pronounced under cold-start or low-activity conditions, where user profiles and interaction histories are limited or unavailable.

Recent deep learning models, including LSTM-based~\cite{lee2019app,xia2020deepapp} and Transformer-based architectures~\cite{kang2022app, zhang2024optimizing}, have shown improvements by modeling sequential dependencies in app usage. Further, methods like AppUsage2Vec~\cite{zhao2019appusage2vec} and DUGN~\cite{ouyang2022learning} incorporate contextual information, such as location or app category, to enrich temporal modeling. Despite these advances, two critical limitations remain unresolved:

\begin{itemize}
    \item {Static or discretized temporal encoding:
Most existing models treat time in a coarse or categorical fashion, using fixed bins (e.g., morning/afternoon/night) or discrete time tokens~\cite{wang2019modeling, khaokaew2024maple, zhang2024optimizing}. This overlooks the fine-grained and continuous nature of temporal user behavior, making it difficult to capture subtle shifts in activity over time.}
    
    \item {Lack of adaptive temporal gating:
Uniform aggregation across time steps prevents the model from identifying critical usage patterns, impairing both interpretability and performance, especially in cold-start scenarios where temporal cues are scarce~\cite{hidasi2015session, xu2020predicting, yang2023atpp}.}
\end{itemize}

These limitations underscore the need for a time-sensitive architecture that can (i) encode temporal information in a continuous and expressive space, and (ii) adaptively condition the learned representation on temporal context. To this end, we propose the TGT model, which integrates learnable sinusoidal time encoding with a lightweight temporal gating mechanism. As illustrated in Figure~\ref{fig:TGT}, TGT jointly models app usage history sequences and temporal context within a unified framework, enabling the network to capture fine-grained and context-aware usage patterns.

Our contributions are as follows:
\begin{itemize}
\item Model Design: We introduce an adaptive temporal gating module that assigns fine-grained importance scores to different hours of the day based on sinusoidal projections, enabling interpretable modeling of user behavior.
\item {Contextualized Feature Encoding: We incorporate heterogeneous nature of user-context data—including session sequence, and temporal—into a unified embedding space, supporting both short-term dependency modeling.}
\item Empirical Validation: We evaluate our model on two real-world datasets under standard and cold-start protocols. TGT consistently outperforms state-of-the-art baselines, particularly in cold-start.
\end{itemize}

\section{Related Work}

App usage prediction is essential for mobile user modeling, supporting proactive services such as app preloading~\cite{yan2012fast}, energy optimization~\cite{neto2021building}, and personalized recommendations~\cite{tian2020cohort}. Existing methods can be grouped into three categories: heuristic and sequence models, graph-based frameworks, and Transformer-based time series predictors.

\subsection{Heuristic and Sequential Models}

Traditional methods like Most Frequently Used (MFU) and Most Recently Used (MRU)~\cite{shin2012understanding} offer simple baselines but neglect temporal and contextual dependencies. To capture sequential dynamics, recurrent models such as LSTM-based DeepApp~\cite{xia2020deepapp}, DeepPattern~\cite{deeppatter}, and NeuSA~\cite{aliannejadi2021context} have been proposed. These models encode app usage along with contextual features (e.g., time and location), but they struggle with long-term dependencies and often rely on user-specific identifiers, limiting generalization to unseen users.

\subsection{Graph-Based and Context-Aware Models}

Graph-based methods such as SA-GCN~\cite{sagcn} and DUGN~\cite{ouyang2022learning} model app co-occurrence or dynamic transitions using graph neural networks. While effective in capturing structural relationships, they often require dense interaction graphs and large datasets, limiting their applicability in lightweight or cold-start scenarios. Semantic-aware models like AppUsage2Vec~\cite{zhao2019appusage2vec} and CoSEM~\cite{khaokaew2021cosem} introduce attention mechanisms or contextual features, but they rely heavily on static app or user embeddings and tend to overlook fine-grained temporal dynamics and context-dependent behavioral variations—factors that are crucial in real-world mobile usage prediction.

\subsection{Transformer-Based and Temporal Models}

Recent works have adopted Transformer architectures to model app usage as temporal sequences. MAPLE~\cite{khaokaew2024maple} integrates multimodal features into a large language model, but it treats time via discrete bins or tokens, failing to capture continuous periodic patterns. Time series models like FEDformer~\cite{zhou2022fedformer}, Reformer~\cite{kitaev2020reformer}, and TimesNet~\cite{wu2022timesnet} offer strong temporal forecasting performance but were not originally designed for user-level behavioral prediction and are less effective in personalized settings.

\section{Methodology}
\subsection{Problem Definition}  

Let $\mathcal{A} = \{a_1, a_2, \dots, a_C\}$ be the set of $C$ unique applications, and let $\mathcal{U} = \{u_1, u_2, \dots, u_N\}$ denote the set of users. Given a mobile app usage dataset, we segment each user's usage history into fixed-length sessions, where each session contains $M$ app usage events without idle gap longer than $\Delta t = 300$ seconds~\cite{pejovic2015anticipatory}.
Each session is denoted as:
\[
\mathbf{S} = \{({A}_m, {T}_m)\}_{m=1}^M,
\]
where:
\begin{itemize}
    \item ${A}_m \in \mathcal{A}$: Identity of the app used at the $m$-th position in session;
    \item ${T}_m \in \mathbb{R}$: Usage duration associated with $\mathbf{A}_m$.
\end{itemize}

In addition, each session is associated with:
\begin{itemize}
    \item $u \in \mathcal{U}$: The user profiles;
    \item $h \in \{0, 1, \dots, 23\}$: The hour of the day when the session ends (temporal context).
\end{itemize}

{Given an observed session $\mathbf{S}$, together with user-specific and temporal contextual attributes $(u, h)$, the goal is to predict the application that the user is most likely to use in the immediate next time step:}
\begin{equation}
    Y = \arg\max_{a_k \in \mathcal{A}} P(a_k \mid \mathbf{S}, u, h),
\end{equation}
where $P(a_k \mid \mathcal{S}, u, h)$ denotes the probability of selecting app $a_k$ given the session and contextual information.

Our model jointly encodes session sequence, user profiles and temporal context to characterize user behavior. %Notably, it does not require user profile at inference time, enabling robust prediction under cold-start and sparse-activity conditions.

\begin{figure*}
    \centering
    \includegraphics[width=\linewidth]{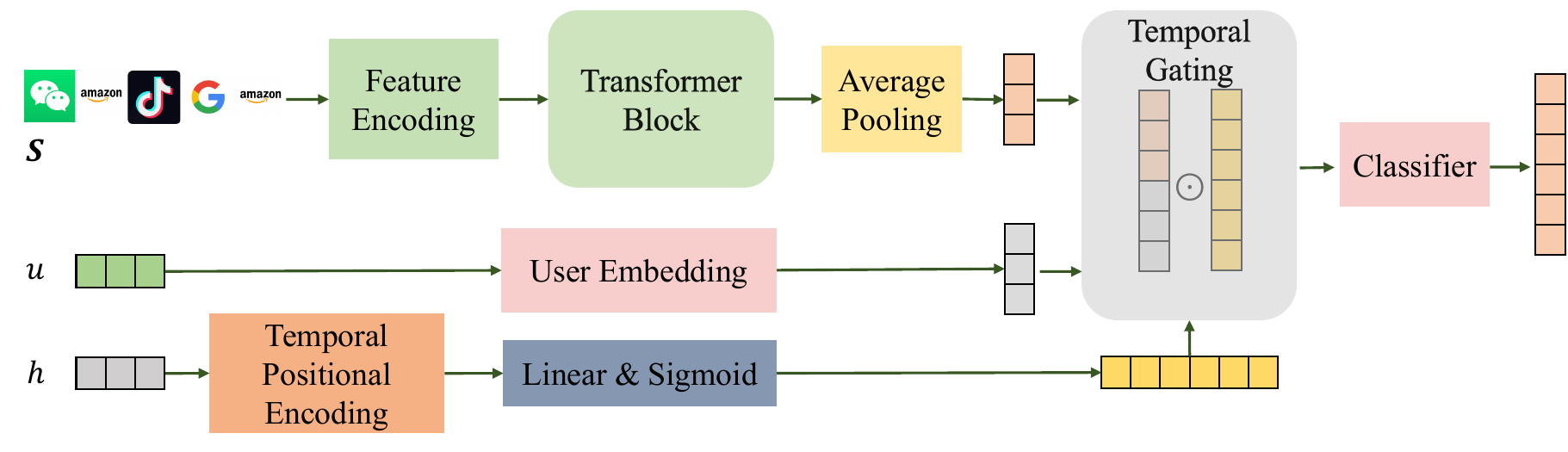}
    \caption{The architecture of the TGT model. The model consists of four main components: 
        (1) \textbf{Feature Encoding}, which extracts app-related representations from historical app sequences; 
        (2) \textbf{Temporal Positional Encoding}, which encodes temporal information to capture usage patterns; 
        (3) \textbf{Transformer Block}, which models contextual dependencies and behavioral dynamics across the sequence; and 
        (4) \textbf{Temporal Gating}, which conditions the pooled representation on time-of-day by adaptively scaling feature dimensions, enabling context-aware modulation of user behavior representations.}
    \label{fig:TGT}
\end{figure*}

\subsection{TGT Architecture}

{We propose TGT, a temporal gating-enhanced Transformer architecture designed to model the dynamic and context-dependent nature of mobile app usage. It jointly encodes session sequences, user profiles and temporal context to support behavior-aware prediction. As illustrated in Figure~\ref{fig:TGT}, the model consists of the following components:}

\begin{itemize}
    \item Feature encoding: Each session sequence is embedded as a series of session vectors, where each vector integrates app identity and usage duration features. 
    
    \item User Embedding: The user profile is separately embedded into a user vector.

    \item {Temporal positional encoding: A Fourier-based positional encoding transforms each time step into a temporal vector using sinusoidal functions at multiple frequencies, injecting temporal context and behavioral dynamics into the sequence.}

    \item {Transformer block: A Transformer encoder models contextual and sequential dependencies across the session vectors.}

    \item {Temporal Gating: The module combines sinusoidal time encoding with a learnable projection to generate a feature-wise gating vector. This vector adaptively scales the dimensions of the pooled session representation via element-wise multiplication, conditioning the representation on temporal context (e.g., time-of-day). This design enables the model to modulate feature importance according to temporal conditions, improving context-aware prediction.}
\end{itemize}

{The aggregated representation is then passed through a classification layer to predict the application that the user is most likely to use at the next time step.}

\subsection{Fourier-Based Encoding}
\label{sec:fourier}
{
To better capture temporal variations in user behavior, we apply sinusoidal transformations—motivated by Fourier analysis—to both usage features and temporal context. Leveraging the expressive capacity of Fourier bases~\cite{carleson1966convergence, fefferman1971convergence, fourier2020}, this encoding enables the model to represent app usage patterns across multiple time scales, supporting fine-grained temporal context modeling.}

\paragraph{Feature Encoding.} Given a session sequence \( \mathbf{S} = \{\mathbf{s}_1, \mathbf{s}_2, \dots, \mathbf{s}_M\} \), where each vector \( \mathbf{s}_m = ({A}_m, {T}_m) \in \mathbb{R}^{2} \), we apply element-wise sinusoidal projections as follows:
\begin{equation}
\bar{\mathbf{s}}_m = \left[ \sin(\mathbf{W}_1 \mathbf{s}_m + \mathbf{b}_1)\parallel\ \cos(\mathbf{W}_2 \mathbf{s}_m + \mathbf{b}_2) \right],
\end{equation}
where \( \mathbf{W}_1, \mathbf{W}_2 \in \mathbb{R}^{d \times 2} \) and \( \mathbf{b}_1, \mathbf{b}_2 \in \mathbb{R}^{d} \) are learnable parameters. This transformation maps each vector into a Fourier-enhanced representation \( \bar{\mathbf{s}}_m \in \mathbb{R}^{2d} \). The full encoded session becomes \( {\bar{\mathbf{S}}} = \{\bar{\mathbf{s}}_1, \dots, \bar{\mathbf{s}}_M\}  \).

\paragraph{Temporal Positional Encoding.}
To encode the hour of day \( h \in \{0, 1, \dots, 23\} \), we adopt a parameter-free sinusoidal encoding:
\begin{equation}
\theta_i = \frac{2\pi(i - h)}{d}, \quad i = 0, 1, \dots, \frac{d}{2} - 1,
\end{equation}
\begin{equation}
\mathbf{P}_h = [\sin(\theta_0), \dots, \sin(\theta_{\frac{d}{2}-1}) \parallel \cos(\theta_0), \dots, \cos(\theta_{\frac{d}{2}-1})] \in \mathbb{R}^d.
\end{equation}
By mapping discrete time steps into a continuous embedding space, this encoding supports robust generalization across different temporal contexts.

\subsection{Transformer Block and User Embedding}

\paragraph{Transformer Block.}

The encoded sequence  \( {\bar{\mathbf{S}}} = \{\bar{\mathbf{s}}_1, \dots, \bar{\mathbf{s}}_M\}  \) is passed through an \( L \)-layer Transformer encoder to capture sequential dependencies:
\begin{equation}
\begin{split}
\hat{\mathbf{s}}_m^{(0)} &= \bar{\mathbf{s}}_m, m \in \{1,...,M\}\\
\tilde{\mathbf{s}}_m^{(l)} &= \text{LayerNorm}(\hat{\mathbf{s}}_m^{(l)} + \text{MHSA}(\hat{\mathbf{s}}_m^{(l)})), \\
\hat{\mathbf{s}}_m^{(l+1)} &= \text{LayerNorm}(\tilde{\mathbf{s}}_m^{(l)} + \text{FFN}(\tilde{\mathbf{s}}_m^{(l)})),
\end{split}
\end{equation}
where \( \text{MHSA} \) denotes multi-head self-attention, and \( \text{FFN} \) is a feed-forward network. {The output of the final Transformer layer is aggregated via pooling:}
\begin{equation}
\mathbf{X}^{(L)} = \frac{1}{M} \sum_{m=1}^{M} \mathbf{W}_L \hat{\mathbf{s}}_m^{(L)} \in \mathbb{R}^{d},
\end{equation}
where $\mathbf{W}_L \in \mathbb{R}^{d\times 2d}$ is a learnable matrix.

\paragraph{User Embedding.}
Each user \( u \in \{1, \dots, N\} \) is associated with a learnable embedding \( \hat{\mathbf{U}} = \mathbf{E}[u] \in \mathbb{R}^{d} \), where \( \mathbf{E} \in \mathbb{R}^{N \times d} \) is a trainable embedding matrix. The final representation is obtained by concatenating the session encoding with the user embedding:
\begin{equation}
\mathbf{X}_\text{app} = \sigma(\mathbf{X}^{(L)} \parallel \hat{\mathbf{U}}),
\end{equation}
where \( \sigma(\cdot) \) denotes a non-linear activation function and \( \parallel \) is the concatenation operator.

\subsection{Temporal Gating}
To incorporate temporal context into user behavior modeling, we adopt a \textit{temporal gating} mechanism rather than step-wise temporal attention. Given the temporal context encoding $\mathbf{P}_h$, we compute a feature-wise gating vector:
\begin{equation}
\mathbf{g}_t = \sigma(\mathbf{W}_h \mathbf{P}_h),
\end{equation}
where $\mathbf{W}_h \in \mathbb{R}^{2d \times d}$ is a learnable projection and $\sigma(\cdot)$ denotes the sigmoid activation. 

This gating vector adaptively scales the dimensions of the pooled session representation:
\begin{equation}
\mathbf{X}_o = \mathbf{X}_\text{app} \odot \mathbf{g}_t,
\end{equation}
where $\odot$ denotes element-wise multiplication. 

Unlike time-aware attention that reweights sequence elements step by step, temporal gating conditions the final representation on contextual time information (e.g., time-of-day), providing a lightweight yet effective way to adapt feature importance. This design improves interpretability by revealing how different feature dimensions are emphasized or suppressed under varying temporal contexts, and we visualize these feature-wise gates in the analysis section.

\subsection{Classifier and Loss Function}
The final representation $\mathbf{X}_o \in \mathbb{R}^{2d \times 1}$ is mapped to class logits via a linear transformation:
\begin{equation}
\hat{\mathbf{Y}} = \mathbf{W}_o \mathbf{X}_o  + \mathbf{b}_o,
\end{equation}
where $\mathbf{W}_o \in \mathbb{R}^{C \times 2d}$ and $\mathbf{b}_o \in \mathbb{R}^C$.

The predicted class probabilities are given by:
\begin{equation}
\hat{P}_{i,c} = \frac{\exp(\hat{\mathbf{Y}}_{i,c})}{\sum_{j=1}^{C} \exp(\hat{\mathbf{Y}}_{i,j})}, \quad c = 1,\dots,C.
\end{equation}

We train the model using the cross-entropy loss:
\begin{equation}
\mathcal{L} = -\frac{1}{N} \sum_{i=1}^{N} \log \hat{P}_{i,y_i},
\end{equation}
where $y_i \in \{1,\dots,C\}$ is the ground-truth class index for the $i$-th sample. During inference, the predicted label is $\hat{y}_i = \arg\max_c \hat{P}_{i,c}$.

\subsection{Key Differences from Transformers}

Unlike prior Transformer-based models for app usage prediction, TGT introduces two key innovations. 
First, it employs {Fourier-based temporal encoding}, replacing coarse time bins with continuous sinusoidal projections to capture fine-grained and periodic user behavior. 
Second, it incorporates a {temporal gating mechanism}, which adaptively scales feature dimensions according to time-of-day, providing lightweight yet interpretable temporal modulation. 
Together, these designs address the limitations of static time encoding and uniform aggregation in standard Transformers, yielding improved prediction and interpretability.

\section{Experiments}

\subsection{Dataset}
We evaluate TGT on two widely-used real-world datasets: Tsinghua App Usage~\cite{yu2018smartphone} and LSapp~\cite{AliannejadiTOIS21}. Key dataset statistics are summarized in Table~\ref{tab:dataset_comparison}.

\begin{table}[htbp]
    \centering
    \caption{Comparison of Tsinghua App Usage and LSapp Datasets}
    \label{tab:dataset_comparison}
    \resizebox{\columnwidth}{!}{ 
    \begin{tabular}{lccccc}
        \toprule
        \textbf{Dataset} & \textbf{Users} & \textbf{Apps} & \textbf{Records} & \textbf{Time Span} & \textbf{Attributes} \\
        \midrule
        \multirow{3}{*}{Tsinghua App Usage} & \multirow{3}{*}{1000} & \multirow{3}{*}{2000 }& \multirow{3}{*}{4,171,950} & \multirow{3}{*}{7 days} & User ID, Timestamp, \\ 
        & & & & & Base Station ID, App ID, \\ 
        & & & & & Interaction Type \\
        \midrule
        \multirow{3}{*}{LSapp} & \multirow{3}{*}{292} &\multirow{3}{*}{ 87} & \multirow{3}{*}{599,635} & \multirow{3}{*}{24 Hours} & User ID, Timestamp, \\ 
        & & & & & Session ID, App ID, \\ 
        & & & & & Interaction Type\\
        \bottomrule
    \end{tabular}}
\end{table}

We follow a unified data preprocessing pipeline, including session segmentation and timestamp normalization. Complete details of data processing are described in Appendix~\ref{appendix:data_pro}.

\subsection{Split Method}
For the two datasets, we adopt two widely used split protocols from MAPLE~\cite{khaokaew2024maple}:

\begin{itemize}
    \item {Standard Setting}: Each user's data is split chronologically into 70\% for training, 10\% for validation, and 20\% for testing. This setup ensures comprehensive model training, fine-tuning, and evaluation.
    
    \item {Cold Start Setting}: To evaluate generalization to unseen users, we adopt a user-level split with 90\% for training and 10\% for testing. User profiles are excluded during both training and inference to prevent information leakage. To ensure fair comparison, the app set is shared across splits, with no unseen apps introduced in testing.
\end{itemize}

\subsection{Baselines}
We compare TGT against a comprehensive set of baselines spanning four modeling paradigms: rule-based heuristics, sequential models, and Transformer-based predictors.

\begin{itemize}
    \item {MFU/MRU}~\cite{shin2012understanding}: Predict the most frequently or most recently used app, respectively.

    \item {AppUsage2Vec}~\cite{zhao2019appusage2vec}, {DeepApp}~\cite{xia2020deepapp}, {NeuSA}~\cite{aliannejadi2021context}: Sequential models using LSTM or attention mechanisms to capture contextual app usage patterns.

    \item {CoSEM}~\cite{khaokaew2021cosem} and {MAPLE}~\cite{khaokaew2024maple}: Models leveraging contextual and semantic embeddings, with MAPLE further incorporating large language model (LLM) embeddings for enhanced app usage prediction.
    \item 
    SparseTSF~\cite{sparsetsf}, Cyclenet~\cite{cyclenet}, TQNet~\cite{lin2025TQNet}: Period-aware models that capture temporal periodicity via frequency sparsity, recurrent cycles, or attention-guided queries.

    \item {Transformer}~\cite{vaswani2017attention}, {Reformer}~\cite{kitaev2020reformer}, {FEDformer}~\cite{zhou2022fedformer}, {TimesNet}~\cite{wu2022timesnet},ContiFormer~\cite{chen2023contiformer}: Recent Transformer variants designed for efficient time series modeling. They were primarily designed for long sequence prediction tasks. In addition, we consider DLinear~\cite{zeng2023transformers} and FreTS~\cite{yi2023frequencydomain}, non-Transformer baselines that employs a simple linear decomposition strategy, providing a strong yet lightweight benchmark for temporal forecasting.
\end{itemize}

To ensure a consistent and meaningful comparison with prior studies while capturing ranking-oriented performance, we adopt Hit Ratio@K (HR@K) and Mean Reciprocal Rank@K (MRR@K) as evaluation metrics. These metrics are widely recognized in temporal prediction and recommendation tasks for quantifying both the accuracy of top-K predictions and the quality of ranking. The formal definitions and justification for their use are provided in Appendix~\ref{app:metrics}. For completeness, Appendix~\ref{app:import_details} summarizes the hyperparameter settings used in our experiments. %and Appendix~\ref{app:code_availability} provides the code repository link to facilitate reproducibility.

\begin{table*}[ht]
\centering
\caption{Comparison of prediction methods under the standard split. Results are reported as mean$_{\pm\mathrm{std}}$ (\%) over 5 runs. Results for NeuSA, CoSEM, and MAPLE are taken from their original papers, where standard deviations were not reported. Best results are in {bold}, second-best are {underlined}.}
 \resizebox{1.0\linewidth}{!}{
\begin{tabular}{lcccccccccc}
\toprule
{\textbf{Datasets} } & \multicolumn{5}{c}{\textbf{Tsinghua App Usage}} & \multicolumn{5}{c}{\textbf{LSapp}} \\
\midrule
\multirow{2}{*}{\textbf{Methods} } & \multicolumn{3}{c}{\textbf{HR@K}} & \multicolumn{2}{c}{\textbf{MRR@K}} & \multicolumn{3}{c}{\textbf{HR@K}} & \multicolumn{2}{c}{\textbf{MRR@K}} \\
\cmidrule(lr){2-4} \cmidrule(lr){5-6} \cmidrule(lr){7-9}  \cmidrule(lr){10-11} 
 & \textbf{1} & \textbf{3} & \textbf{5} & \textbf{3}  & \textbf{5}  & \textbf{1} & \textbf{3} & \textbf{5} & \textbf{3}  & \textbf{5} \\
\midrule
MFU
& 14.60$_{\pm 0.7}$ & 27.72$_{\pm 0.6}$ & 35.33$_{\pm 0.8}$ & 20.23$_{\pm 1.0}$ & 21.95$_{\pm 0.4}$ 
& 22.17$_{\pm 0.2}$ & 34.81$_{\pm 0.4}$ & 43.65$_{\pm 0.2}$ & 32.80$_{\pm 0.3}$ & 42.08$_{\pm 0.1}$ \\
MRU
& 8.21$_{\pm 0.1}$ & 25.05$_{\pm 0.9}$ & 37.23$_{\pm 0.7}$ & 15.30$_{\pm 0.1}$ & 18.08$_{\pm 0.1}$ 
& 13.92$_{\pm 0.2}$ & 18.70$_{\pm 0.6}$ & 23.89$_{\pm 0.1}$ & 18.73$_{\pm 0.1}$ & 36.79$_{\pm 0.8}$ \\
Transformer
& 49.46$_{\pm 0.1}$ & 67.18$_{\pm 0.8}$ & 73.31$_{\pm 0.4}$ & 57.36$_{\pm 0.6}$ & 58.76$_{\pm 0.3}$ 
& 80.95$_{\pm 0.1}$ & 91.63$_{\pm 0.4}$ & 93.42$_{\pm 0.1}$ & 87.06$_{\pm 0.7}$ & 87.47$_{\pm 0.6}$ \\
Appusage2Vec 
& 49.33$_{\pm0.3}$ & 69.63$_{\pm0.4}$ & 77.16$_{\pm0.3}$ & 58.34$_{\pm0.2}$ & 60.07$_{\pm0.3}$ 
& 80.28$_{\pm0.2}$ & 94.12$_{\pm0.1}$ & 96.19$_{\pm0.1}$ & \textbf{87.86}$_{\pm0.1}$ & \underline{88.04}$_{\pm0.2}$ \\
Reformer
& 49.51$_{\pm0.4}$ & 67.34$_{\pm0.3}$ & 73.47$_{\pm0.3}$ & 57.45$_{\pm0.3}$ & 58.85$_{\pm0.2}$ 
& 80.95$_{\pm0.2}$ & 91.73$_{\pm0.1}$ & 93.55$_{\pm0.3}$ & 87.09$_{\pm0.1}$ & 87.51$_{\pm0.1}$ \\
DeepApp
& 31.51$_{\pm0.5}$ & 48.63$_{\pm0.6}$ & 55.49$_{\pm0.5}$ & 38.98$_{\pm0.4}$ & 40.56$_{\pm0.2}$ 
& 40.50$_{\pm0.5}$ & 67.41$_{\pm0.4}$ & 78.53$_{\pm0.3}$ & 52.15$_{\pm0.4}$ & 54.71$_{\pm0.1}$ \\
TimesNet 
& \underline{52.13}$_{\pm0.2}$ & 73.47$_{\pm0.2}$ & 80.48$_{\pm0.2}$ & \underline{61.94}$_{\pm0.2}$ & 63.02$_{\pm0.2}$ 
& \underline{81.02}$_{\pm0.1}$ & \underline{93.71}$_{\pm0.1}$ & \underline{95.14}$_{\pm0.1}$ & 83.38$_{\pm0.1}$ & 87.12$_{\pm0.2}$ \\
FEDformer
& 51.37$_{\pm0.2}$ & 73.06$_{\pm0.3}$ & 79.94$_{\pm0.2}$ & 61.03$_{\pm0.2}$ & 63.57$_{\pm0.2}$ 
& 80.41$_{\pm0.2}$ & 93.10$_{\pm0.1}$ & 94.71$_{\pm0.1}$ & 86.30$_{\pm0.1}$ & 87.40$_{\pm0.1}$ \\
DLinear 
& 49.51$_{\pm 0.3}$ & 67.19$_{\pm0.3}$ & 73.36$_{\pm0.3}$ & 57.40$_{\pm 0.2}$ & 58.82$_{\pm0.2}$ 
& 80.96$_{\pm0.2}$ & 91.68$_{\pm0.1}$ & 93.55$_{\pm0.1}$ & 85.09$_{\pm0.1}$ & 87.52$_{\pm0.1}$ \\
FreTS
&50.09$_{\pm0.3}$&65.62$_{\pm0.2}$&70.41$_{\pm0.1}$& 57.06$_{\pm0.3}$ & 58.16$_{\pm0.2}$ & 67.94$_{\pm0.1}$&75.53$_{\pm0.1}$&78.16$_{\pm0.3}$& 59.62$_{\pm0.1}$ & 68.64$_{\pm0.2}$\\
ContiFormer& 34.93 $_{\pm0.1}$& 45.10$_{\pm0.1}$& 51.72$_{\pm0.3}$& 39.68$_{\pm0.2}$& 41.06$_{\pm0.1}$& 64.33 $_{\pm0.1}$& 75.09$_{\pm0.1}$& 81.23$_{\pm0.1}$& 69.81$_{\pm0.1}$& 73.07$_{\pm0.1}$\\
NeuSA
& 46.40 & 65.62 & 72.86 & 54.92 & 56.58
& 68.32 & 82.53 & 88.30 & 74.61 & 75.93 \\
CoSEM
& 41.63 & 66.82 & 74.99 & 52.82 & 54.69
& 49.90 & 74.66 & 81.49 & 60.83& 62.42 \\
MAPLE
& 51.91 & \textbf{73.85} & \textbf{81.15}& 61.69 & 63.38 
& 71.57 & 86.49 & 91.50 & 78.21 & 79.36\\
SparseTSF
& 24.32$_{\pm0.6}$ & 36.72$_{\pm0.6}$ & 40.72$_{\pm0.2}$ & 29.68$_{\pm0.5}$ & 30.59$_{\pm0.4}$ 
& 62.44$_{\pm0.4}$ & 75.92$_{\pm0.3}$ & 81.83$_{\pm0.3}$ & 71.63$_{\pm0.3}$ & 72.56$_{\pm0.2}$ \\
Cyclenet
& 27.80$_{\pm0.5}$ & 45.28$_{\pm0.8}$ & 56.82$_{\pm0.5}$ & 38.47$_{\pm0.4}$ & 42.21$_{\pm0.2}$ 
& 65.28$_{\pm0.3}$ & 76.67$_{\pm0.3}$ & 83.51$_{\pm0.3}$ & 73.18$_{\pm0.3}$ & 76.60$_{\pm0.2}$ \\
TQNet
& 24.87$_{\pm0.5}$ & 37.38$_{\pm0.2}$ & 41.83$_{\pm0.6}$ & 30.31$_{\pm0.4}$ & 31.34$_{\pm0.4}$ 
& 65.78$_{\pm0.3}$ & 77.94$_{\pm0.3}$ & 83.89$_{\pm0.2}$ & 71.15$_{\pm0.3}$ & 73.63$_{\pm0.3}$ \\
\midrule
\textbf{TGT} 
& \textbf{52.66}$_{\pm 0.1}$ 
& \underline{73.70}$_{\pm 0.2}$ 
& \underline{80.50}$_{\pm 0.0}$ 
& \textbf{62.11}$_{\pm 0.1}$ 
& \textbf{63.67}$_{\pm 0.1}$ 
& \textbf{81.15}$_{\pm 0.1}$ 
& \textbf{94.94}$_{\pm 0.1}$ 
& \textbf{96.67}$_{\pm 0.2}$ 
& \underline{87.67}$_{\pm 0.1}$ 
& \textbf{88.07}$_{\pm 0.1}$ \\
\bottomrule
\end{tabular}
}
\label{tab:standard_split_results}
\end{table*}

\begin{table*}[ht]
\centering
\caption{Comparison of prediction methods under the cold-start split. Results are reported as percentages mean$_{\pm\mathrm{std}}$ (\%) over 5 runs. Best results are in bold; second-best are underlined.}
\label{tab:cold_start_results}
 \resizebox{1.0\linewidth}{!}{
\begin{tabular}{lcccccccccc}
\toprule
{\textbf{Datasets} } & \multicolumn{5}{c}{\textbf{Tsinghua App Usage}} & \multicolumn{5}{c}{\textbf{LSapp}} \\
\midrule
\multirow{2}{*}{\textbf{Methods} } & \multicolumn{3}{c}{\textbf{HR@K}} & \multicolumn{2}{c}{\textbf{MRR@K}} & \multicolumn{3}{c}{\textbf{HR@K}} & \multicolumn{2}{c}{\textbf{MRR@K}} \\
\cmidrule(lr){2-4} \cmidrule(lr){5-6} \cmidrule(lr){7-9}  \cmidrule(lr){10-11} 
 & \textbf{1} & \textbf{3} & \textbf{5} & \textbf{3}  & \textbf{5}  & \textbf{1} & \textbf{3} & \textbf{5} & \textbf{3}  & \textbf{5} \\
\midrule
MFU
& 4.89$_{\pm0.2}$ & 15.49$_{\pm0.5}$ & 25.35$_{\pm0.4}$ & 8.57$_{\pm0.1}$ & 10.97$_{\pm0.3}$ 
& 1.55$_{\pm0.3}$ & 3.73$_{\pm0.5}$ & 8.71$_{\pm0.2}$ & 1.27$_{\pm0.3}$ & 2.31$_{\pm0.2}$ \\
MRU
& 9.11$_{\pm0.3}$ & 28.59$_{\pm0.4}$ & 43.82$_{\pm0.5}$ & 17.28$_{\pm0.4}$ & 20.76$_{\pm0.5}$ 
& 18.25$_{\pm0.1}$ & 26.25$_{\pm0.7}$ & 34.61$_{\pm0.2}$ & 21.28$_{\pm0.5}$ & 29.30$_{\pm0.3}$ \\
Transformer
& 61.56$_{\pm0.3}$ & 78.10$_{\pm0.3}$ & 82.80$_{\pm0.3}$ & 68.88$_{\pm0.2}$ & 69.95$_{\pm0.2}$ 
& 81.49$_{\pm0.2}$ & 93.61$_{\pm0.1}$ & 94.88$_{\pm0.1}$ & 86.62$_{\pm0.1}$ & 87.55$_{\pm0.1}$ \\
Reformer
& 60.73$_{\pm0.3}$ & 77.90$_{\pm0.3}$ & 82.69$_{\pm0.3}$ & 68.38$_{\pm0.3}$ & 69.47$_{\pm0.2}$ 
& \underline{81.70}$_{\pm0.2}$ & 93.56$_{\pm0.1}$ & 94.96$_{\pm0.1}$ & 82.88$_{\pm0.2}$ & 86.12$_{\pm0.1}$ \\
TimesNet
& 63.79$_{\pm0.2}$ & \underline{83.97}$_{\pm0.3}$ & \underline{88.24}$_{\pm0.2}$ & \underline{73.04}$_{\pm0.2}$ & \underline{74.02}$_{\pm0.2}$ 
& \underline{81.70}$_{\pm0.1}$ & \underline{95.19}$_{\pm0.2}$ & 95.77$_{\pm0.1}$ & 82.89$_{\pm0.1}$ & 85.61$_{\pm0.1}$ \\
FEDformer
& \underline{63.93}$_{\pm0.2}$ & 82.63$_{\pm0.2}$ & 87.13$_{\pm0.2}$ & 72.95$_{\pm0.2}$ & 73.98$_{\pm0.1}$ 
& 81.36$_{\pm0.1}$ & 94.93$_{\pm0.1}$ & \underline{96.30}$_{\pm0.1}$ & \underline{87.19}$_{\pm0.1}$ & \underline{88.26}$_{\pm0.1}$ \\
DLinear
& 61.63$_{\pm0.3}$ & 77.49$_{\pm0.3}$ & 82.36$_{\pm0.3}$ & 68.64$_{\pm0.3}$ & 69.75$_{\pm0.2}$ 
& 81.39$_{\pm0.2}$ & 93.62$_{\pm0.1}$ & 95.00$_{\pm0.1}$ & 82.63$_{\pm0.1}$ & 85.82$_{\pm0.1}$ \\
FreTS
&57.92$_{\pm0.2}$&69.23$_{\pm0.1}$&77.11$_{\pm0.1}$& 54.37$_{\pm0.3}$ & 66.28$_{\pm0.1}$ & 83.92$_{\pm0.3}$&85.44$_{\pm0.2}$&88.50$_{\pm0.2}$& 67.24$_{\pm0.1}$ & 79.33$_{\pm0.1}$\\
ContiFormer
& 56.30 $_{\pm0.1}$& 66.02$_{\pm0.1}$& 77.24$_{\pm0.1}$& 58.84$_{\pm0.1}$& 71.67$_{\pm0.1}$& 77.29 $_{\pm0.1}$& 85.95$_{\pm0.1}$& 92.36$_{\pm0.1}$& 78.14$_{\pm0.1}$& 82.74$_{\pm0.1}$\\
NeuSA
& 44.33 & 61.69 & 68.12 & 52.06 & 53.53 
& 68.74& 77.70 & 81.35 & 72.72 & 73.55 \\
CoSEM
& 31.11 & 55.97 & 65.25 & 42.04 & 44.16
& 45.23 & 72.43& 81.04 & 57.18 & 59.18 \\
MAPLE
& 52.28 & 74.17 & 81.28 & 62.06 & 63.69 
& 76.44 & 88.48 & 92.47 & 81.81 & 82.72 \\
SparseTSF
& 42.44$_{\pm0.1}$ & 66.72$_{\pm0.1}$ & 71.29$_{\pm0.2}$ & 60.58$_{\pm0.1}$ & 63.87$_{\pm0.1}$ 
& 62.31$_{\pm0.2}$ & 72.25$_{\pm0.2}$ & 78.37$_{\pm0.3}$ & 66.34$_{\pm0.1}$ & 75.62$_{\pm0.4}$ \\
Cyclenet
& 47.03$_{\pm0.1}$ & 68.84$_{\pm0.1}$ & 72.67$_{\pm0.7}$ & 64.57$_{\pm0.6}$ & 71.20$_{\pm0.3}$ 
& 65.28$_{\pm0.4}$ & 74.65$_{\pm0.2}$ & 81.14$_{\pm0.3}$ & 71.31$_{\pm0.1}$ & 77.06$_{\pm0.3}$ \\
TQNet
& 43.70$_{\pm0.1}$ & 67.88$_{\pm0.1}$ & 71.60$_{\pm0.1}$ & 61.09$_{\pm0.3}$ & 64.36$_{\pm0.2}$ 
& 63.88$_{\pm0.5}$ & 73.19$_{\pm0.4}$ & 78.95$_{\pm0.2}$ & 65.11$_{\pm0.4}$ & 76.34$_{\pm0.2}$ \\
\midrule
\textbf{TGT} 
& \textbf{64.25$_{\pm 0.2}$}
& \textbf{84.45$_{\pm 0.1}$ }
& \textbf{89.11$_{\pm 0.2}$} 
& \textbf{73.42$_{\pm 0.2}$ }
& \textbf{74.49$_{\pm 0.2}$ }
& \textbf{82.02$_{\pm 0.1}$ }
& \textbf{95.46$_{\pm 0.1}$} 
& \textbf{96.71$_{\pm 0.1}$ }
& \textbf{88.47$_{\pm 0.1}$ }
& \textbf{88.76$_{\pm 0.1}$} \\
\bottomrule
\end{tabular}
}
\end{table*}

\subsection{Results on Standard Split}

On the standard splits of the Tsinghua and LSapp datasets (Table~\ref{tab:standard_split_results}), TGT surpasses all baselines in HR@1 on Tsinghua and achieves competitive performance on LSapp, confirming its consistent generalization ability.

Unlike prior approaches such as AppUsage2Vec, DeepApp, and NeuSA, which model usage sequences or user profiles in isolation, TGT jointly encodes both modalities through a Transformer backbone augmented with a temporal gating mechanism. This design allows the model to dynamically condition on both app usage and temporal context, enabling behavior-aware prediction. Compared to generic time-series forecasting models like TimesNet and DLinear—which treat session sequences as homogeneous feature series and are typically optimized for long-range forecasting—TGT is more suitable for short-horizon, time-sensitive prediction tasks where session-level dependencies and temporal context play a critical role. We also observe that periodic modeling baselines such as Cyclenet, TQNet, and SparseTSF consistently underperform Transformer-based architectures on both datasets. This performance gap can be attributed to their reliance on strong global periodicity assumptions, which often break down in real-world scenarios characterized by irregular, noisy, and context-dependent user routines. In contrast, TGT employs a temporal gating mechanism that adaptively identifies and emphasizes informative time steps, enabling robust and context-adaptive modeling beyond fixed cyclic patterns.

\subsection{Results on Cold Start Split}
We evaluate model performance under the user cold-start setting, where test users are entirely unseen during training. To prevent information leakage, all models are trained without user profile or embeddings, and only compliant baselines are included. As shown in Table~\ref{tab:cold_start_results}, TGT surprisingly outperforms all baselines across all metrics, despite not using user-specific features. This seemingly counterintuitive result can be attributed to four key factors:

\begin{itemize}
    \item {More diverse training data:}  
   {Compared to the standard setting (70\% training users), the cold-start protocol uses 90\% of users for training, exposing the model to a broader range of usage patterns. Although evaluation users are unseen, this larger training set helps the model learn more generalizable temporal dynamics.}

    \item {User-agnostic robustness:}  
    {TGT captures transferable behavioral patterns through session modeling and temporal gating, without relying on user profile. This design mitigates overfitting to individual users and emphasizes population-level regularities, enhancing generalization.}

    \item {Adaptability to variable temporal context:}  
    {Unlike time-series baselines such as FEDformer and DLinear, which assume stationarity and rely on long-range continuity—TGT dynamically attends to salient temporal subsequences, making it more robust under sparse and context-varying cold-start conditions.}

    \item {Limitations of periodic models:}  
    {Periodic models like Cyclenet, TQNet, and SparseTSF underperform in the cold-start setting due to their reliance on fixed-frequency assumptions. These assumptions break down for users with irregular or non-stationary routines. In contrast, TGT does not assume global periodicity and adaptively attends to informative time contexts, enabling stronger generalization.}
\end{itemize}

\subsection{Running Time Comparison}

\begin{figure}[ht]
\centering
\includegraphics[width=1.0\linewidth]{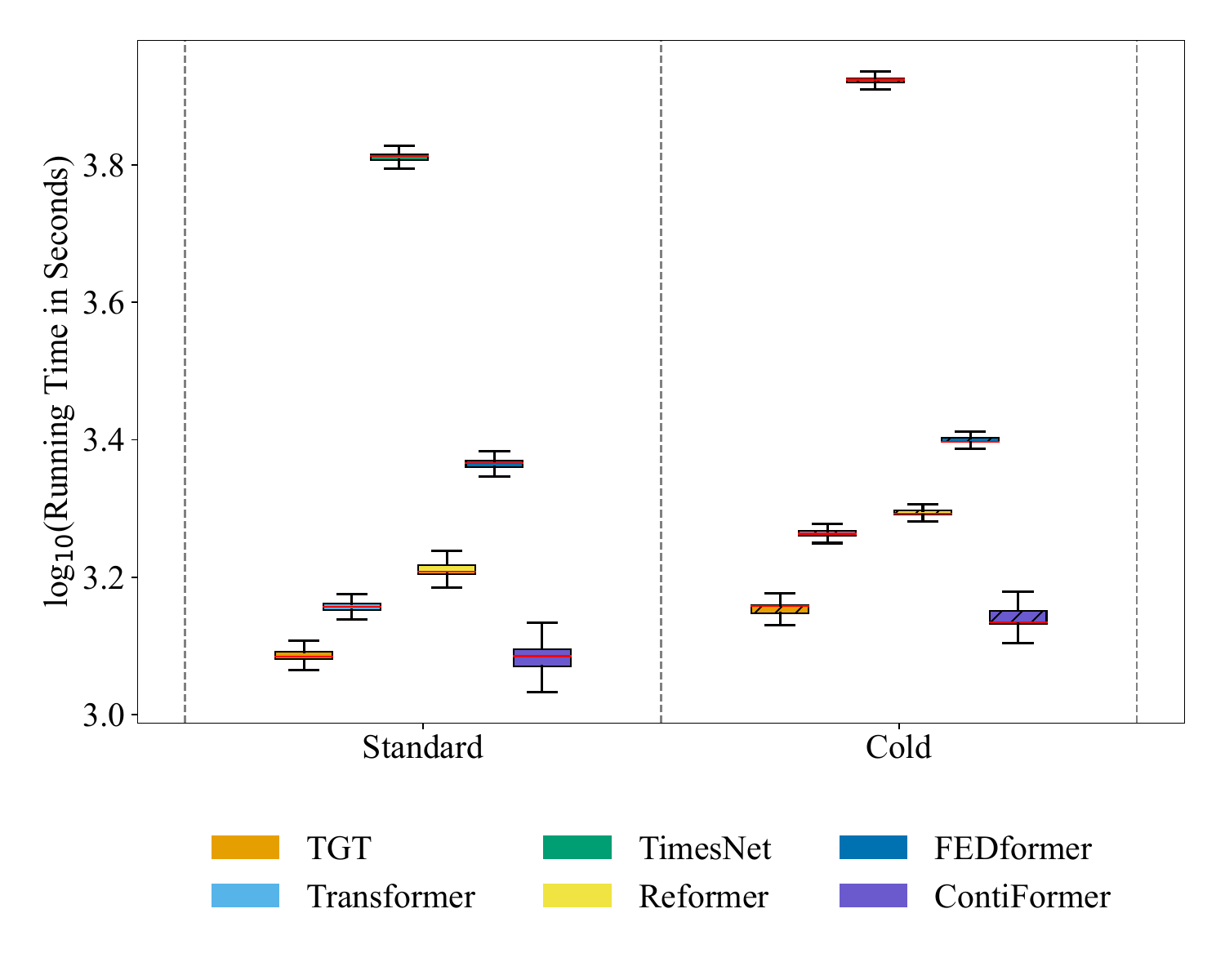}
\caption{Boxplot of $\log_{10}$-scaled training time (in seconds) for five models on the Tsinghua App Usage dataset, evaluated under both Standard and Cold-start settings. Each result is averaged over five runs of 10 epochs.}
\label{fig:runtimes}
\end{figure}

We compare the training time of five representative models—TGT, Transformer, TimesNet, Reformer, and FEDformer—on the Tsinghua App Usage dataset under both \textit{Standard} and \textit{Cold-start} settings. All experiments are conducted on identical hardware over five runs of 10 epochs each.

{As shown in Figure~\ref{fig:runtimes}, TimesNet incurs the highest runtime due to its deep, multi-scale structure. Reformer and FEDformer show moderate cost, benefiting from efficient attention or spectral operations. TGT and the vanilla {Transformer} are the most efficient, owing to simpler architectures. }

\subsection{Ablation Study}

We conduct two sets of ablation studies on the Tsinghua App Usage dataset (standard split) to assess the contribution of key components in TGT: {(1) input features, and (2) feature encoding strategies for feature encoding and temporal positional encoding.} All results are averaged over five independent runs, with standard deviation as error bars.

\begin{figure}[ht]
    \centering    \includegraphics[width=1.0\linewidth]{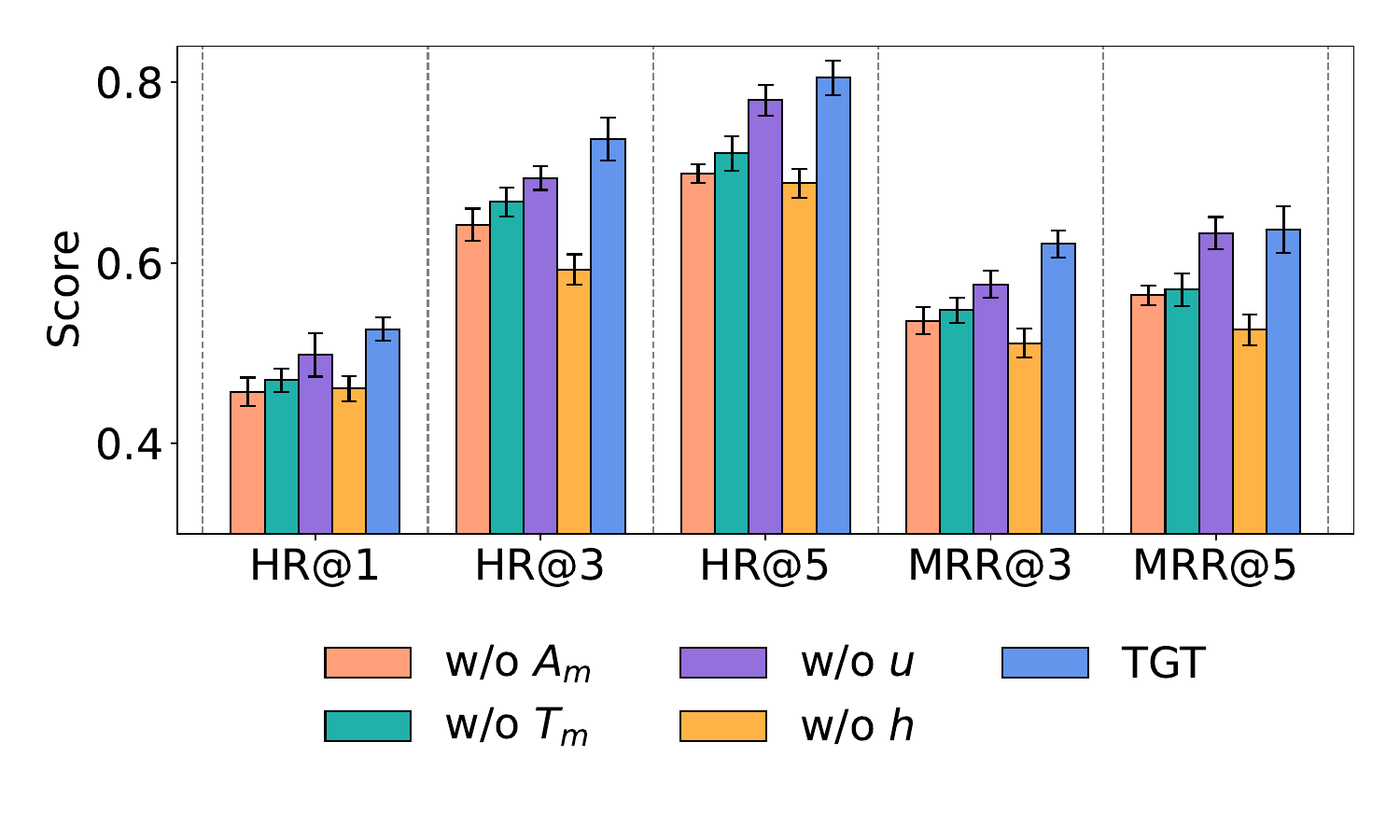}
    \caption{
    Input feature ablation results of TGT on the Tsinghua App Usage dataset. Each group of bars shows the average HR@K and MRR@K scores over five runs, with error bars indicating the standard deviation.
    }
    \label{fig:ablation}
\end{figure}

\paragraph{Feature Ablation Analysis.}
Figure~\ref{fig:ablation} summarizes the impact of removing individual input features. Each feature—app identify, usage duration, user profile, and temporal gating—encodes distinct behavioral cues that contribute jointly to model performance.

\begin{itemize}
    \item {{Without app identity ($w/o~{A}_m$):}  
Removing app identifiers from the usage history significantly degrades performance, underscoring the importance of sequential context in capturing user intent. Without past app information, the model loses temporal anchors necessary for predicting the next action.}

\item {{Without usage duration ($w/o ~{T}_m$):}  
Excluding usage duration impairs the model’s ability to capture recency effects and session-level temporal dependencies. Time intervals and usage lengths are critical for distinguishing deliberate interactions from incidental or transient usage.}

\item {{Without user profile ($w/o~u$):}  
Removing user embeddings leads to moderate performance degradation, suggesting that while temporal gating captures shared behavioral patterns, personalized user signals still contribute to improved top-K prediction accuracy.}

\item {Without temporal gating ($w/o~h$):
Disabling the temporal gating module causes an 18.58\% drop in HR@1, showing that the Transformer encoder with self-attention alone cannot fully account for temporal context. Temporal gating complements self-attention by conditioning the pooled representation on time-of-day, adaptively reweighting feature dimensions to capture context-specific behavioral patterns.}

\end{itemize}

\begin{figure}[ht]
    \centering
    \includegraphics[width=1.0\linewidth]{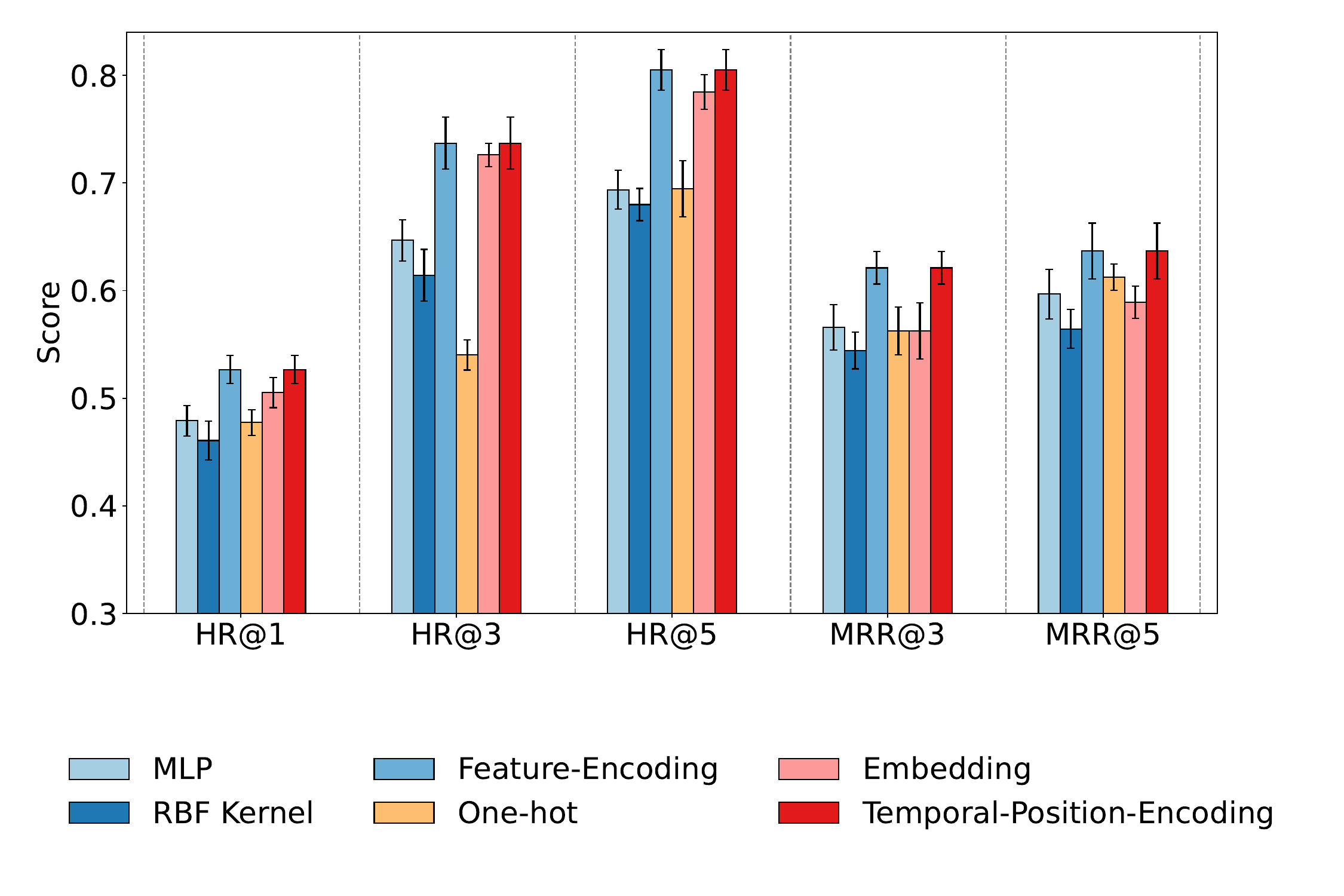}
    \caption{
        Comparison of feature encoding and temporal positional encoding on the Tsinghua App Usage dataset. Each bar indicates average HR@K or MRR@K over five runs; error bars show standard deviation.
    }
    \label{fig:feature}
\end{figure}

\paragraph{Encoding Strategy Ablation.}
We further evaluate the effect of Fourier-based encoding on both feature encoding and temporal positional encoding, as shown in Figure~\ref{fig:feature}. 

\begin{itemize}
   \item {Feature encoding.} Compared with MLP and RBF kernel baselines, the Fourier projection consistently achieves better HR@K and MRR@K scores. By applying sinusoidal transformations with learnable linear mappings, the encoding captures correlations between app identity and usage duration and introduces diverse periodic patterns.
    
   \item {Temporal positional encoding.} The sinusoidal encoding maps discrete hour indices into a continuous embedding space, which improves prediction consistency across adjacent hours. Compared with one-hot or learnable embeddings, it better reflects gradual changes in user behavior over time. 
\end{itemize}

Overall, these results indicate that Fourier-based encodings provide more stable and generalizable representations than conventional alternatives, particularly under sparse or irregular usage patterns.

\subsection{Training-Size Effects under Cold-Start}

\begin{figure}[ht]
\centering
\includegraphics[width=1.0\linewidth]{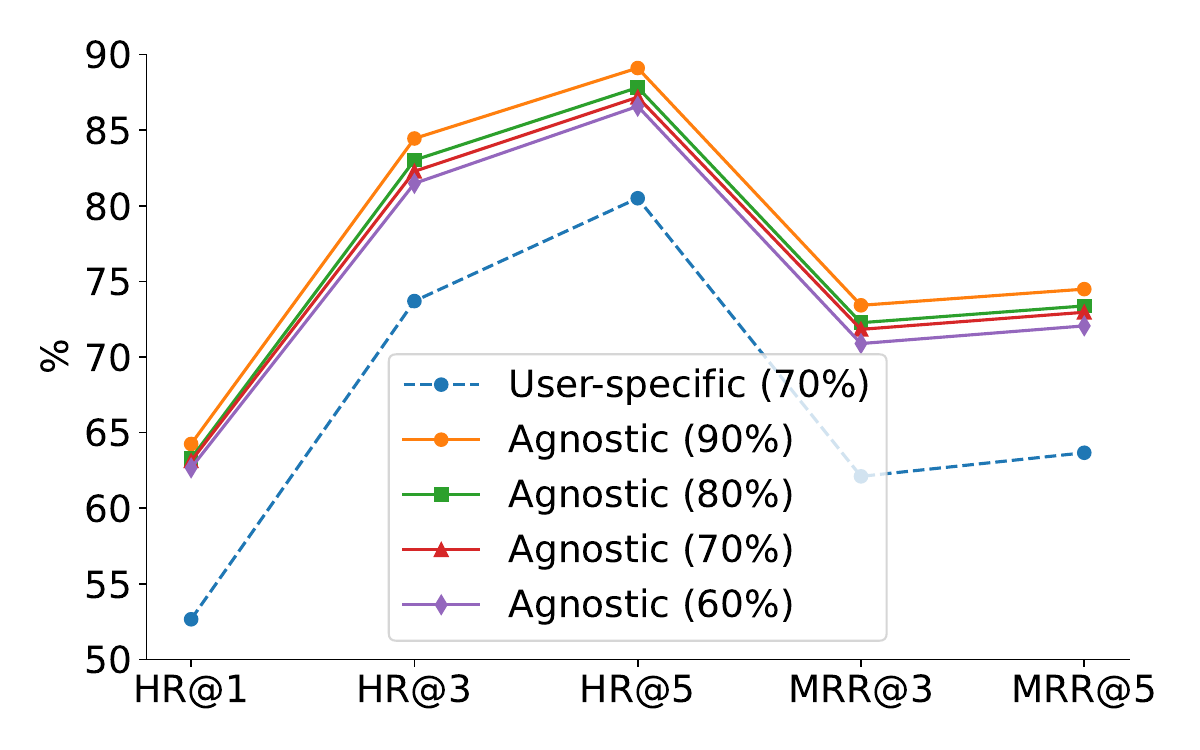}
\caption{Performance of user-agnostic models under different training user proportions (90\%, 80\%, 70\%, 60\%). 
The dashed line indicates the user-specific baseline trained with 70\% of users.}
\label{fig:coldstart_size}
\end{figure}

To disentangle the effect of training data size from the benefit of user-agnostic modeling, we vary the proportion of training users while fixing the test set. Figure~\ref{fig:coldstart_size} visualizes the results of user-agnostic models trained with 90\%, 80\%, 70\%, and 60\% of users, along with the user-specific baseline at 70\%. As expected, performance gradually declines when fewer users are available, confirming that larger training sets naturally provide more signal.

Crucially, when restricted to the same proportion of training users (70\%) as in the user-specific baseline, the user-agnostic model still achieves substantially higher HR@K and MRR@K scores. This demonstrates that the observed gains cannot be solely attributed to training size. Instead, they stem from the inductive bias of parameter sharing across users, which enables more effective generalization to unseen or sparsely observed users in the cold-start regime.

It is important to note that the weaker performance of user-specific models in this figure does not imply that user information is useless. In warm-start scenarios where test users also appear in training, user embeddings provide clear benefits. However, in the cold-start setting, user-specific models fail to generalize because their user ID embeddings cannot be pre-trained for unseen users. In contrast, user-agnostic models avoid this issue by learning cross-user patterns that transfer to new users.

\subsection{Interpretability Analysis of Temporal Gating}

To evaluate the interpretability of the temporal gating mechanism, we visualize the hour-conditioned gating weights learned by TGT on the Tsinghua App Usage dataset (standard split). For each hour of the day, we average the feature-wise gates $\mathbf{g}_t$ across sessions, revealing how temporal context modulates the relative importance of different representation dimensions.

\begin{figure}[ht]
\centering
\includegraphics[width=1.0\linewidth]{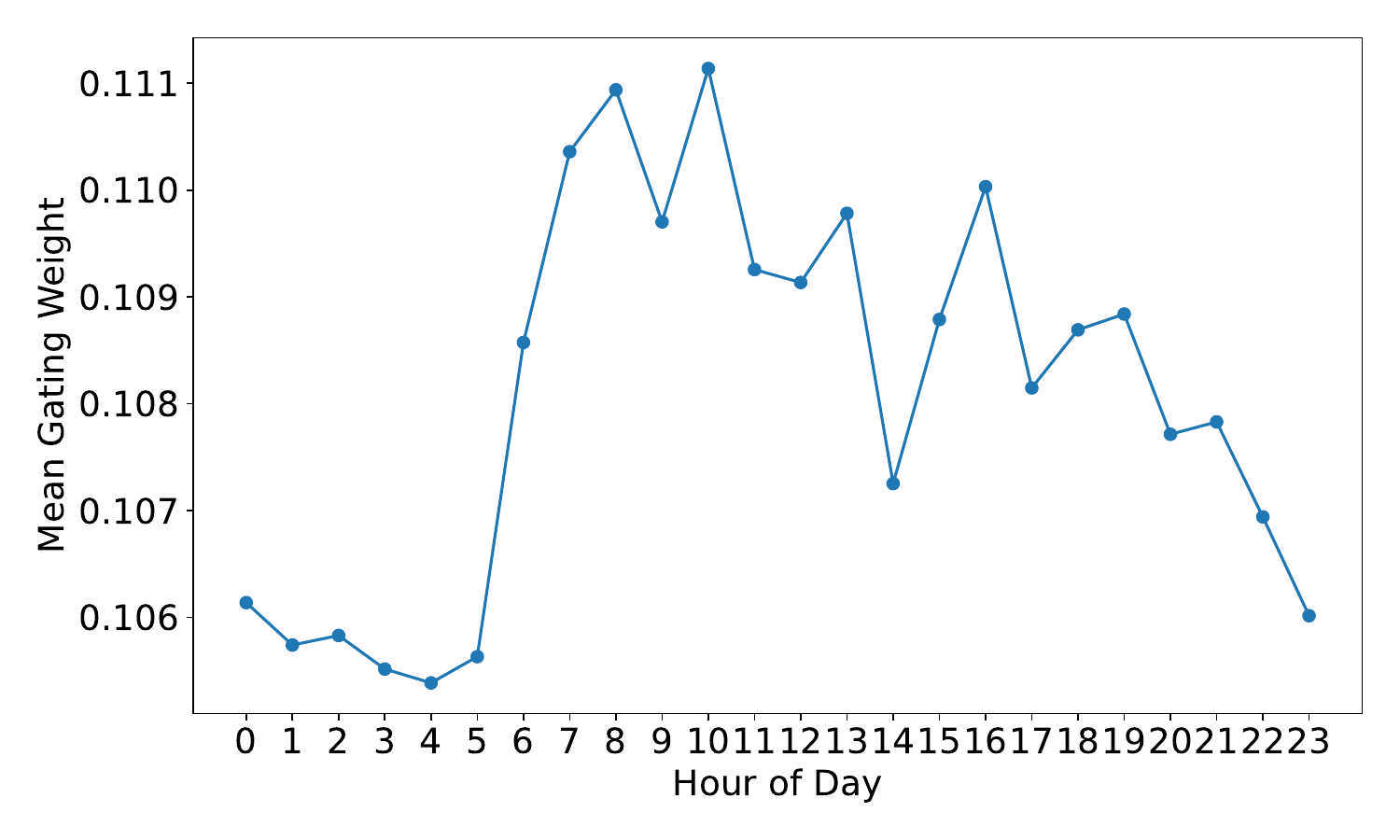}
\caption{Hour-conditioned average gating values learned by TGT on the Tsinghua App Usage dataset. Temporal context adaptively scales feature dimensions, capturing time-dependent behavioral patterns.}
\label{fig:time_curve}
\end{figure}

Figure~\ref{fig:time_curve} shows that the model assigns different importance levels to features depending on temporal context. Gating values are higher during high-usage periods such as mid-morning (10 a.m.) and late afternoon (4 p.m.), while early morning hours (1–5 a.m.) exhibit consistently lower scaling factors. This analysis confirms that the gating mechanism does not collapse to a global bias but instead captures dynamic, hour-specific modulation of feature dimensions, enabling the model to represent time-aware user behavior.

\subsection{Additional Experiments}
In addition to the above experiments, TGT also conducted four other experiments: Hyperparameter Sensitivity~\ref{app:hyper_sen},  Training Convergence~\ref{app:train_conv}, Time Threshold Sensitivity~\ref{app:time_th}, Time-based Split~\ref{app:time_split}, Comparison of Model Variants~\ref{app:com_model}, and Case Study~\ref{app:case}.

\section{Additional Experimental}
In addition to the above experiments, I also conducted four other experiments, namely Hyperparameter Sensitivity, Time Threshold Sensitivity, Time-based Split, Comparison of Model Variants, and Case Study: Visualizing Top-K Predictions with Category Mapping.

\section{Conclusion}
In this work, we proposed TGT, a Transformer architecture enhanced with a temporal gating mechanism for modeling sparse and irregular app usage behavior. TGT conditions feature representations on temporal context, complementing self-attention and enabling more accurate and interpretable predictions. Extensive experiments demonstrated consistent improvements over competitive baselines, particularly under cold-start scenarios where user identity information is unavailable. These results highlight the effectiveness of incorporating temporal modulation as an inductive bias for user behavior modeling.

Looking ahead, we plan to investigate lightweight Transformer variants for efficient on-device deployment, and to extend TGT towards personalized prediction by integrating user-specific preferences with cross-user transferability. Another promising direction is a deeper analysis of temporal gating to provide finer-grained interpretability of how temporal context shapes behavioral dynamics.

\section*{Acknowledgments}
This work was supported by the National Natural Science Foundation of China [Grants 12471488].

\appendices

\section{Processing Details}

\subsection{Data Processing}
\label{appendix:data_pro}
To ensure consistency in sequence generation, both datasets undergo a preprocessing phase, including:

\begin{itemize}
    \item Filtering inactive users and apps: We exclude users with fewer than 50 app usage records and apps with fewer than 10 occurrences to ensure data reliability.
    
    \item {App Usage Session}: We define app usage sessions by grouping consecutive app usage events within a time threshold of $\Delta t $=300 seconds. This threshold balances session continuity while minimizing noise from long idle periods. 

    \item {Timestamp normalization}: All timestamps are converted to a uniform format using Coordinated Universal Time (UTC) to preserve temporal consistency.
\end{itemize}

This preprocessing ensures the generated sequences maintain temporal coherence and enhance the model's ability to predict future app usage effectively.

\subsection{Evaluation Metrics} \label{app:metrics}
To evaluate the effectiveness of our next mobile app recommendation model, we compare the recommended top-K list with the ground truth. We primarily evaluate models using Hit Ratio@K (HR@K) and Mean Reciprocal Rank@K (MRR@K), which are widely adopted in prior work. 

HR@K is calculated as the ratio of hits in the top-K recommendations to the total number of test samples, defined as:
\[
\text{HR@K} = \frac{\#\text{hit@K}}{|N_{\text{test}}|},
\]
where \(\#\text{hit@K}\) represents the number of hits in the test dataset and \(|N_{\text{test}}|\) signifies the test dataset’s total quantity.

MRR@K emphasizes correct predictions at higher ranks by computing the reciprocal of the rank of the first correct recommendation:
\[
\text{MRR@K} = \frac{1}{|N_{\text{test}}|} \sum_{i=1}^{|N_{\text{test}}|} \frac{1}{\text{rank}_i},
\]
where \(\text{rank}_i\) is the rank of the $i$-th correct prediction.

NDCG@K accounts for the position of hits within the recommendation list, giving higher scores to hits at higher ranks. It is calculated as:
\[
\text{NDCG@K} = \frac{\sum_{i=1}^K \frac{2^{\text{rel}_i} - 1}{\log_2(i+1)}}{\sum_{j=1}^{|REL_K|} \frac{2^{\text{rel}_j} - 1}{\log_2(j+1)}},
\]
where \(\text{rel}_i\) denotes the graded relevance of the result at position i, and \(|REL_K|\) is the number of predictions in the result ranking list up to position K.

For all metrics, larger values indicate better prediction performance, capturing both accuracy and ranking quality.

\subsection{Implementation Details}
\label{app:import_details}
Our experiments were conducted on an NVIDIA RTX 3090 GPU, with 24GB RAM and an AMD Ryzen Threadripper PRO 5975WX 32-Cores CPU. The model was implemented using PyTorch 2.2.2 with CUDA 12.1 support. 

\begin{table}[ht]
    \centering
    \caption{Hyperparameter settings}
    \begin{tabular}{lc}
        \toprule
        \textbf{Hyperparameter} & \textbf{Value} \\
        \midrule
        Batch size & 512 \\
        Learning rate & 0.001 \\
        Optimizer & Adam \\
        Hidden size & 128 \\
        Dropout rate & 0.2 \\
        Number of Transformer layers & 2 \\
        Number heads& 4\\
        Training epochs & 50 \\
        Early stopping patience & 5 \\
        \bottomrule
    \end{tabular}
        
    \label{tab:detail}
\end{table}

To prevent overfitting and ensure efficient training, we employed early stopping with a patience of 5 epochs, meaning that training was terminated if the test loss did not improve for five consecutive epochs. The hyperparameter settings can befound in Table~\ref{tab:detail}. All hyperparameters were selected via grid search on the validation set.

\iffalse
\subsection{Code Availability}
\label{app:code_availability}
The implementations of the models used in this work are publicly available:  
\begin{itemize}
    \item \textbf{TGT}:~\url{https://github.com/LongLee220/Atten-Transformer}.
\item\textbf{DeepApp}:~\url{https://github.com/anonymous1833/DeepApp}
\item\textbf{DLinear}:~\url{https://github.com/cure-lab/LTSF-Linear}
\item\textbf{AppUsage2Vec}:~\url{https://github.com/zzaebok/AppUsage2Vec}. 
    \item \textbf{FEDformer}:~\url{https://github.com/MAZiqing/FEDformer}.
    \item \textbf{Reformer}:~\url{https://github.com/lucidrains/reformer-pytorch}.
    \item \textbf{TimesNet}:~\url{https://github.com/thuml/TimesNet}.
    \item \textbf{Transformer}:~\url{https://github.com/bkhanal-11/transformers}
    
    \item \textbf{FreTS}:~\url{https://github.com/aikunyi/FreTS?tab=readme-ov-file}
    \item 
    \textbf{ContiFormer}~\url{https://github.com/microsoft/SeqML/tree/main/ContiFormer}
\end{itemize}

These repositories provide source code, training procedures, and additional implementation details for reproducibility.
\fi

\section{Additional Experimental}
\label{appendix:add_experient}

All experiments in this appendix are conducted on the Tsinghua App Usage dataset under the standard split setting. The goal is to further analyze model behavior in terms of robustness, design components, and interpretability.

\subsection{Hyperparameter Sensitivity}  
\label{app:hyper_sen}
We conduct a systematic investigation into three key hyperparameters—Transformer depth, input sequence length, and dropout rate—to understand their influence on the model’s temporal modeling capacity and generalization behavior. Rather than focusing on absolute accuracy values, we interpret how each setting affects the learning dynamics of TGT.

\begin{table}[ht]
\centering
\caption{Performance comparison of different hyperparameters.}
\begin{tabular}{lccccc}
\toprule
\multirow{2}{*}{\textbf{Types}} & \multicolumn{3}{c}{\textbf{HR@K}}  & \multicolumn{2}{c}{\textbf{MRR@K}} \\
\cmidrule(lr){2-4} \cmidrule(lr){5-6}
 & \textbf{1} & \textbf{3} & \textbf{5} & \textbf{3}  & \textbf{5} \\
\midrule
\multicolumn{6}{l}{\textbf{Transformer Layers}} \\
\midrule
1  & 0.4970 & 0.7032 & 0.7875 & 0.5642 & 0.5825 \\
2  &\textbf{0.5266} & \textbf{0.7370} & \textbf{0.8050} & \textbf{0.6211} & \textbf{0.6367} \\
3  & 0.5021 & 0.7213 & 0.7952 & 0.5367 & 0.5763 \\
4  & 0.5123 & 0.6907 & 0.7637 & 0.5992 & 0.6125  \\
\midrule
\multicolumn{6}{l}{\textbf{Sequence Length}} \\
\midrule
3  & 0.4854 & 0.6972 & 0.7617 & 0.5759 & 0.5961   \\
4  & 0.5010 & 0.7142 & 0.7684 & 0.5980 & 0.6174  \\
5  &0.5266 & \textbf{0.7370} & \textbf{0.8050} & 0.6211 & 0.6367 \\
6  & 0.5257 & 0.7345 & 0.8024 & 0.6014 & 0.6171  \\ 
7  & {0.5302} & 0.72781 & 0.7886 & 0.5966 & 0.6350  \\
8  &\textbf{0.5372} & {0.7310} & {0.8028} & \textbf{0.6318} & \textbf{0.6559}  \\
\midrule
\multicolumn{6}{l}{\textbf{Dropout Probability}} \\
\midrule
0.0  & 0.5014 & 0.6723 & 0.7864 & 0.5473 &  0.5764 \\
0.1  & 0.5007 & 0.6848 & 0.7981 & 0.5686 & 0.5954  \\
0.2  &\textbf{0.5266} & \textbf{0.7370} & \textbf{0.8050} & \textbf{0.6211} & \textbf{0.6367}  \\
0.3  & 0.5157 & 0.6886 & 0.7861 & 0.5685 & 0.5861  \\
0.4  & 0.4921 & 0.6434 & 0.6974 & 0.5238 & 0.5408   \\ 
0.5  & 0.4272 & 0.6342 & 0.6804 & 0.5146 & 0.5377   \\ 
0.6  & 0.4183 & 0.6190 & 0.6702 & 0.5035 & 0.5249   \\ 
\bottomrule
\end{tabular}

\label{tab:hyperparams}
\end{table}

From Table~\ref{tab:hyperparams}, we observe:
\begin{itemize}
    \item Transformer Depth: Increasing the number of layers from 1 to 2 significantly improves performance, suggesting that a two-layer attention stack suffices to capture short- to mid-range dependencies. Further depth leads to diminishing returns or overfitting, indicating limited long-range temporal gain in this domain.
    \item Sequence Length: Model performance improves as the sequence length increases from 3 to 5 or 6, then plateaus. This supports the intuition that recent behaviors dominate user intent, while overly long sequences introduce noise. Slight gains in ranking metrics (e.g., MRR) with longer windows suggest benefits for capturing broader behavioral patterns.
    \item Dropout Rate: Moderate dropout (e.g., 0.2) improves generalization by mitigating overfitting to skewed patterns and temporal noise. Excessive dropout harms performance, especially in sparse sessions, by discarding subtle behavioral cues. A balanced regularization strategy is thus essential.

\end{itemize}

\subsection{Training Convergence under Standard Split}
\label{app:train_conv}

\begin{figure*}[ht]
    \centering
    \includegraphics[width=1.0\linewidth]{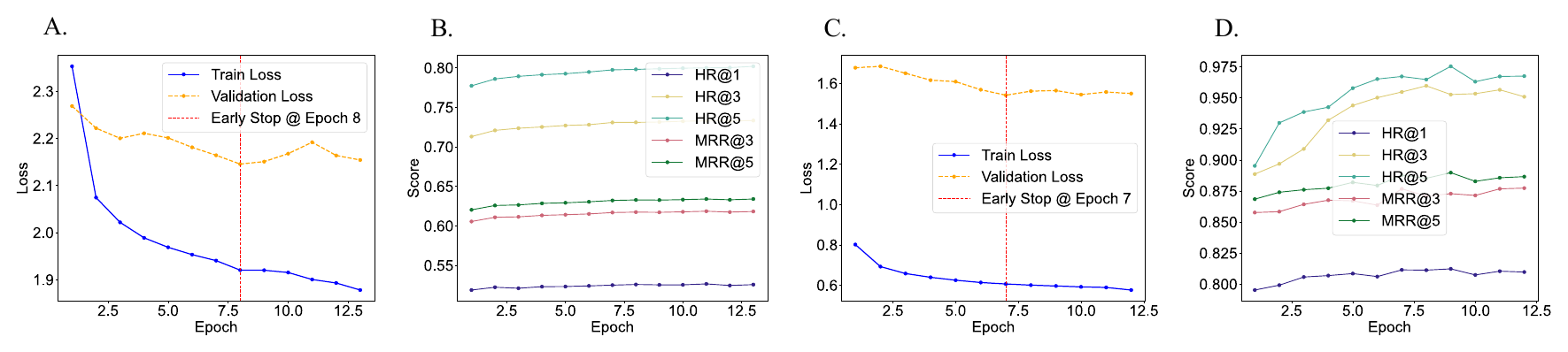}
    \caption{Training convergence of TGT on the Tsinghua App Usage dataset and LSApp dataset under the standard split. 
    A. Training and validation loss curves for the Tsinghua App Usage dataset. 
    B. HR@K and MRR@K evolution on the test set during training on the Tsinghua App Usage dataset. 
    C. Training and validation loss curves for the LSApp dataset. 
    D. HR@K and MRR@K evolution on the test set during training on the LSApp dataset.}
    \label{fig:epoch}
\end{figure*}

Figure~\ref{fig:epoch} illustrates the training convergence of TGT on the Tsinghua and LSapp datasets under the standard split. As shown in Figure~\ref{fig:epoch} A and C, both training and validation loss decrease steadily without divergence, indicating stable convergence and no evident overfitting. Figure~\ref{fig:epoch} B and D further show that HR@K and MRR@K metrics continue to improve throughout training, suggesting that the model not only fits well but also generalizes effectively. These results confirm that TGT achieves consistent optimization and strong generalization across datasets.

\subsection{Time Threshold Sensitivity}
\label{app:time_th}

\begin{table}[ht]
\centering
\caption{Performance comparison of different time thresholds.}
\begin{tabular}{lccccc}
\toprule
\multirow{2}{*}{\textbf{$\Delta t$ (s)}} & \multicolumn{3}{c}{\textbf{HR@K}} & \multicolumn{2}{c}{\textbf{MRR@K}} \\
\cmidrule(lr){2-4} \cmidrule(lr){5-6}
& \textbf{1} & \textbf{3} & \textbf{5} & \textbf{3} & \textbf{5} \\
\midrule
420 s  & 0.4958 & 0.7129 & 0.7608 & 0.5883 & 0.6007  \\   
360 s  & 0.5193 & 0.7271 & 0.7915 & 0.5999 & 0.6197  \\
250 s  & 0.5074 & 0.7184 & 0.7659 & 0.5863 & 0.5960  \\
200 s  & 0.4839 & 0.7041 & 0.7476 & 0.5450 & 0.5810  \\
\midrule
300 s  &\textbf{0.5266} & \textbf{0.7370} & \textbf{0.8050} & \textbf{0.6211} & \textbf{0.6367} \\
\bottomrule
\end{tabular}

\label{tab:time}
\end{table}

Session segmentation controls the granularity of user activity sequences, directly impacting model performance. We vary the idle gap threshold $ \Delta t \in \{200, 250, 300, 360, 420\}$ (Table~\ref{tab:time}) and observe that $\Delta t = 300\,\text{s}$ consistently achieves the best results. This threshold balances capturing intent-driven app bursts with avoiding fragmented or overly aggregated sessions, both of which degrade temporal coherence. Smartphone usage typically consists of short, context-dependent interactions, so this moderate segmentation preserves meaningful patterns while maintaining discriminative power for prediction.

\subsection{Time-based Split}
\label{app:time_split}

For the Tsinghua App Usage dataset, in addition to the standard and cold-start splits, we further adopt a \textit{time-based split} following DUGN~\cite{ouyang2022learning} and Appformer~\cite{sun2025appformer}. Specifically, app usage sessions from the first six consecutive days are used for training, while the sessions from the seventh day constitute the test set. This setting better reflects real-world deployment scenarios where models must generalize from past historical behaviors to predict future app usage. 

As shown in Table~\ref{tab:th_time_split_results}, Atten-Transformer significantly outperforms all baseline methods under this time-based evaluation protocol. Compared with the strongest baseline Appformer, our model achieves a 49.95\% improvement in HR@1, a 27.34\% increase in HR@5, and a 39.77\% boost in MRR@5. These substantial gains highlight the effectiveness of Atten-Transformer in capturing long-term temporal dependencies and its superior generalization ability in realistic temporal recommendation scenarios.

\begin{table*}[htbp]
\centering
\caption{Performance comparison of different recommendation systems on the Tsinghua App Usage dataset with time-based split (first six days for training and the seventh day for testing).}
\label{tab:th_time_split_results}
\resizebox{\textwidth}{!}{ 
\begin{tabular}{lccccccccccccccc}
\toprule
\multirow{2}{*}{\textbf{Methods} }& \multicolumn{5}{c}{\textbf{HR@K}} &\multicolumn{5}{c}{\textbf{NDCG@K}} & \multicolumn{5}{c}{\textbf{MRR@K}} \\
\cmidrule(lr){2-6} \cmidrule(lr){7-11} \cmidrule(lr){12-16}
 & \textbf{1} & \textbf{2} & \textbf{3}  & \textbf{4} & \textbf{5} & \textbf{1} & \textbf{2} & \textbf{3}  & \textbf{4} & \textbf{5} & \textbf{1} & \textbf{2} & \textbf{3}  & \textbf{4} & \textbf{5} \\
\midrule
MRU            & 0.2398 & 0.4196 & 0.5135 & 0.5670 & 0.5938 & 0.2398 &0.3501&0.3921&0.4163&0.4299 & 0.2398 & 0.3210 & 0.3522 & 0.3668 & 0.3754 \\
MFU            & 0.2472 & 0.4205 & 0.5703 & 0.5621 & 0.5987 &0.2472&0.3565&0.3999&0.4235&0.4377 & 0.2472 & 0.3338 & 0.3628 & 0.3765 & 0.3838\\
BPRMF          & 0.3293 & 0.4437 & 0.5188 & 0.5681 & 0.6077&0.3293&0.4015&0.4390&0.4602&0.4755 & 0.3293 & 0.3865 & 0.4115 & 0.4238 & 0.4317 \\
GRU3Rec& 0.3137 & 0.4493 & 0.5425& 0.6028& 0.6494&0.3137&0.3993&0.4459&0.4827&0.5012 & 0.3137 & 0.3815 & 0.4126 & 0.4277 & 0.4370 \\
AppUsage2Vec   & 0.3333 & 0.4592 & 0.5436 & 0.6080 & 0.6560&0.3333 & 0.4127 & 0.4549 & 0.4827& 0.5012& 0.3333& 0.3962 & 0.4244& 0.4405& 0.4501\\
SR-GNN         & 0.3342 & 0.4716 & 0.5563 & 0.6154 & 0.6626&0.3342&0.4209&0.4632&0.4887&0.5070 & 0.3342 & 0.4029 & 0.4311 & 0.4459 & 0.4554 \\
DUGN           & 0.3479 & 0.4768 & 0.5593 & 0.6215 & 0.6710&0.3479&0.4292&0.4705&0.4973&0.5164 & 0.3479 & 0.4124 & 0.4399 & 0.4554& 0.4653 \\
Appformer& 0.4268 &0.5550 &0.6230 &0.6656& 0.6960&0.4268&0.5550&0.5979&0.6192&0.6323 & 0.4268& 0.4909 & 0.5136 & 0.5242 & 0.5303\\
\midrule
\textbf{Atten-Transformer}  & \textbf{0.6400} & \textbf{0.7852} & \textbf{0.8367} & \textbf{0.8663} & \textbf{0.8863} & 
\textbf{0.6400}  & \textbf{0.7316} & \textbf{0.7574} & \textbf{0.7701} & \textbf{0.7779} & 
\textbf{0.6400} & \textbf{0.7126} & \textbf{0.7230} & \textbf{0.7372} & \textbf{0.7412}\\
\bottomrule
\end{tabular}}
\end{table*}

\begin{table}[ht]
\centering
\caption{Performance comparison of different model components.}
\begin{tabular}{lccccc}
\toprule
\multirow{2}{*}{\textbf{Methods}} & \multicolumn{3}{c}{\textbf{HR@K}} &\multicolumn{2}{c}{\textbf{MRR@K}} \\
\cmidrule(lr){2-4} \cmidrule(lr){5-6}
 & \textbf{1} & \textbf{3} & \textbf{5} & \textbf{3}  & \textbf{5}  \\
\midrule
Atten-GRU     &  0.5053 & 0.7306 & 0.8007 & 0.6219 & 0.6315  \\
Atten-RNN     & 0.4833 & 0.7284 & 0.7961 & 0.6193 & 0.6244  \\
Atten-LSTM    & 0.5199 & 0.7351 & 0.7995 & 0.6200 & 0.6376  \\
Atten-CNN     & 0.4802 & 0.6250 & 0.7414 & 0.5134 & 0.5311  \\
Atten-MLP     & 0.4921 & 0.6436 & 0.7607 & 0.5312 & 0.5507  \\
Atten-Random  & 0.4120 & 0.5954 & 0.6530 & 0.4342 & 0.5250  \\
\midrule
\textbf{TGT}  
&\textbf{0.5266} & \textbf{0.7370} & \textbf{0.8050} & \textbf{0.6211} & \textbf{0.6367} \\
\bottomrule
\end{tabular}
\label{tab:module}
\end{table}

\begin{table*}[ht]
\centering
\caption{
Case study of 12 test users showing predicted Top-5 categories and Jaccard similarity with respect to recently used app categories. The true category is also listed for each user.
}
\begin{tabular}{lcccc}
\toprule
{\textbf{User}} & \textbf{Input category} & \textbf{Output category} & \textbf{ Jaccard similarity}& \textbf{True category}\\
\midrule
\multirow{2}{*}{0} & \multirow{2}{*}{Social, Travel}& \textbf{Social}, Travel& \multirow{2}{*}{0.50} & \multirow{2}{*}{Social}\\
 & & News, Utilities& & \\
 
\multirow{2}{*}{2} & Lifestyle, Navigation, News,& \textbf{News}, Lifestyle& \multirow{2}{*}{0.50} & \multirow{2}{*}{News} \\
 &Photo\&Video, Travel&Navigation, Utilities  & &  \\
 
\multirow{2}{*}{5} & \multirow{2}{*}{Entertainment, Social, Utilities} & \textbf{Social}, Travel& \multirow{2}{*}{0.67} & \multirow{2}{*}{Social}\\
 & &News, Utilities & &  \\

\multirow{2}{*}{6} & \multirow{2}{*}{Music, Utilities}& Music, \textbf{Utilities} & \multirow{2}{*}{0.50} & \multirow{2}{*}{Utilities}\\
 & &News, Utilities & &  \\

\multirow{2}{*}{8} & \multirow{2}{*}{News, Social}& \textbf{Social}, Travel& \multirow{2}{*}{0.50} & \multirow{2}{*}{Social}\\
 & & Social, Navigation& &  \\

\multirow{2}{*}{19} & \multirow{2}{*}{Lifestyle, Social, Utilities}& \textbf{Social}, Lifestyle
&\multirow{2}{*}{0.75}& \multirow{2}{*}{Social}\\
 & & Utilities, Navigation& &  \\

\multirow{2}{*}{37}& \multirow{2}{*}{Navigation, Social}& \textbf{Navigation}, Weather&\multirow{2}{*}{0.67}&\multirow{2}{*}{Navigation}\\
 & & Social, Navigation& &  \\

\multirow{2}{*}{41}&\multirow{2}{*}{Lifestyle, Social}&\textbf{Social}, Lifestyle&\multirow{2}{*}{0.50}&\multirow{2}{*}{Social}\\
 & & Utilities, Books& &  \\

\multirow{2}{*}{80}&\multirow{2}{*}{Navigation, Utilities}&Navigation, \textbf{Utilities} &\multirow{2}{*}{0.67} & \multirow{2}{*}{Utilities}\\
 & & Lifestyle& &  \\

\multirow{2}{*}{109}&\multirow{2}{*}{Navigation, News, Utilities}&\textbf{Utilities}, Navigation& \multirow{2}{*}{1.00}&\multirow{2}{*}{Utilities}\\
 & & News& &  \\

\multirow{2}{*}{156}& \multirow{2}{*}{News, Social, Travel}& Social, News& \multirow{2}{*}{0.50}&\multirow{2}{*}{Social}\\
 & & Travel, Finance & &  \\

\multirow{2}{*}{255}& \multirow{2}{*}{Navigation, Social, Utilities}& \textbf{Navigation},  Utilities & \multirow{2}{*}{1.00} &\multirow{2}{*}{Navigation}\\
 & & Social& &  \\
 
\bottomrule
\end{tabular}
\label{tab:case}
\end{table*}

\subsection{Comparison of Model Variants}
\label{app:com_model}

To isolate the contribution of self-attention in TGT, we compare it with several representative architectures, including RNN-based (GRU~\cite{chung2014empirical}, LSTM~\cite{hochreiter1997long}), convolutional~\cite{lecun2002gradient} and feed-forward models~\cite{rosenblatt1958perceptron}. We also include a random-attention variant to examine the necessity of learned attention scores. Results are shown in Table~\ref{tab:module}.

\begin{itemize}
    \item Self-Attention vs. Sequence Encoders. Compared to Atten-GRU and Atten-LSTM, TGT achieves consistently higher HR and MRR scores, particularly at top-1. This reflects the ability of self-attention to directly model long-range dependencies without relying on hidden state propagation. While RNNs effectively capture local patterns, their reliance on sequential recurrence introduces gradient vanishing and makes distant signals harder to retain. In contrast, the parallelizable attention mechanism attends directly to relevant time steps, yielding more expressive representations.
    \item Self-Attention vs. CNN. CNNs focus on local windows and learn hierarchical patterns through stacked layers, but lack a global view over the sequence. The lower performance of Atten-CNN suggests that app usage prediction—driven by both recent context and periodic recurrence—requires flexible attention spans that CNNs cannot easily replicate.
    \item Self-Attention vs. MLP. Replacing attention with MLP (Atten-MLP) removes the model’s ability to selectively focus on relevant past interactions. The resulting performance degradation highlights that attention-based selection, not just representation capacity, is crucial for identifying influential historical signals.
    \item Learned Attention vs. Random Weights. The Atten-Random variant eliminates learned attention by replacing attention scores with fixed random values. Its dramatic performance drop confirms that the effectiveness of TGT stems from learned attention distributions that adaptively weigh context based on temporal relevance. This result rules out the possibility that the improvements arise merely from increased model depth or parameter count.
\end{itemize}

Together, these comparisons demonstrate that self-attention is not just a powerful encoding mechanism, but one that aligns naturally with the irregular, multi-scale temporal structure of app usage data. Its flexibility in weighting historical interactions, independence from positional bias, and compatibility with periodic encoding make it uniquely suited for modeling user behavior sequences.

\subsection{Case Study: Semantic Consistency of Top-K Predictions}
\label{app:case}

To qualitatively assess the model's Top-K prediction performance and its semantic consistency with recent user behavior, we conduct a case study on seven representative users from the test set. For each user, we examine whether the ground-truth app appears in the model’s Top-5 predictions. To improve interpretability, all app IDs are mapped to high-level semantic categories using the dataset-provided \texttt{App2Category} dictionary.

{As shown in Table~\ref{tab:case}, the model correctly includes the target app category in the Top-5 predictions in most cases. To further evaluate the semantic alignment between predictions and recent activity, we compute the Jaccard similarity~\cite{jaccard1912distribution} between the predicted Top-5 categories and the set of input categories observed in the user’s recent session. The resulting scores, ranging from 0.50 to 1.00, indicate a meaningful degree of category overlap. This suggests that TGT not only predicts the correct app in most cases but also produces semantically coherent outputs that reflect the user’s recent behavioral context.}

\bibliographystyle{IEEEtran}
\bibliography{references}

\end{document}